\documentclass[letterpaper]{article}

\def\arxiv{} 

\usepackage{uai2019}
\usepackage[margin=1in]{geometry}

\usepackage{times}
\usepackage{microtype,graphicx,subcaption,booktabs,soul,amsmath,bm,amsthm,xcolor,chapterbib,pifont}
\usepackage{hyperref}
\hypersetup{
    urlbordercolor={white},
}
\usepackage{cleveref}
\usepackage{xspace}
\usepackage[ruled,vlined]{algorithm2e}
\Crefname{algocf}{Algorithm}{Algorithms}

\newtheorem{theorem}{Theorem}

\newtheorem{corollary}[theorem]{Corollary}
\newtheorem{remark}{Remark}[theorem]

\DeclareMathOperator*{\argmin}{arg\,min}

\DeclareMathOperator*{\diag}{\mathrm{diag}}

\newcommand{\kl}{\mathrm{KL}}
\newcommand{\fim}{\mathcal{G}}
\newcommand{\trace}{\mathrm{tr}}
\newcommand{\dx}{\mathrm{d}X}

\makeatletter
\DeclareRobustCommand\onedot{\futurelet\@let@token\bmv@onedotaux}
\def\bmv@onedotaux{\ifx\@let@token.\else.\null\fi\xspace}
\def\eg{\emph{e.g}\onedot} 
\def\ie{\emph{i.e}\onedot}

\def\wrt{w.r.t\onedot}

\makeatother

\begin{document}

\title{Fisher-Bures Adversary Graph Convolutional Networks}
\author{{\bf{}Ke Sun$^\dagger$\thanks{~~Corresponding author}%
~~~~~~~~Piotr Koniusz$^\dagger{}^\ddagger$~~~~~~~Zhen Wang$^\dagger$}\\
$^\dagger$CSIRO Data61\hspace{2em}$^\ddagger$Australia National University\\%
\{Ke.Sun,~Peter.Koniusz,~Jeff.Wang\}@data61.csiro.au}
\maketitle

\begin{abstract}
In a graph convolutional network, we assume that the graph $G$ is generated \wrt some observation noise.
During learning, we make small random perturbations $\Delta{}G$ of the graph and try to improve generalization.
Based on quantum information geometry,
$\Delta{}G$ can be characterized by the eigendecomposition of the graph Laplacian matrix.
We try to minimize the loss \wrt the perturbed $G+\Delta{G}$ while making
$\Delta{G}$ to be effective in terms of the Fisher information of the neural network.
Our proposed model can consistently improve graph convolutional networks on
semi-supervised node classification tasks with reasonable computational overhead.
We present three different geometries on the manifold of graphs:
the intrinsic geometry measures the information theoretic dynamics of a graph;
the extrinsic geometry characterizes how such dynamics can affect externally a graph neural network;
the embedding geometry is for measuring node embeddings.
These new analytical tools are useful in developing a good understanding of
graph neural networks and fostering new techniques.
\end{abstract}

\section{INTRODUCTION}\label{sec:intro}

Recently, neural network architectures are introduced~\cite{gmANM,sgTGN,bzSNA,dbCNN,kwSSC,hyIRL,vcGAN}
to learn high level features of objects based on a given graph among these objects.
These graph neural networks, especially graph convolutional networks (GCNs), showed record-breaking scores on diverse learning tasks.
Similar to the idea of data augmentation, this paper improves GCN generalization
by minimizing the expected loss \wrt small random perturbations of the input graph.
In order to do so, we must first have a rigorous definition of the \emph{manifold of graphs} denoted by $\mathcal{M}$,
which is the space of all graphs satisfying certain constraints.
Then, based on the local geometry of $\mathcal{M}$ around a graph $G\in\mathcal{M}$,
we can derive a compact parameterization of the perturbation so that it can be plugged into a GCN.
We will show empirically that the performance of GCN can be improved
and present theoretical insights on the differential geometry of $\mathcal{M}$.


\subsection*{Notations}

We assume an undirected graph $G$ without self-loops consisting of $n$ nodes indexed as $1,\cdots,n$.
${X}_{n\times{D}}$ denotes the given node features,
${H}_{n\times{d}}$ denotes some learned high-level features,
and ${Y}_{n\times{O}}$ denotes the one-hot node labels.
All these matrices contain one sample per row.
The graph structure is represented by the adjacency matrix $A_{n\times{n}}$ that can be binary or weighted,
so that $a_{ij}\ge0$, and $a_{ij}=0$ indicates no link between nodes $i$ and $j$.
The neural networks weights are denoted by the matrix $W^l$, where $l=1,\cdots,L$ indexes the layers.
We use capital letters such as $A$, $B$, $\cdots$ to denote matrices
and small letters such as $a$, $b$, $\cdots$ to denote vectors.
We try to use Greek letters such as $\alpha$, $\beta$, $\cdots$ to denote scalars.
These rules have exceptions.

\subsection*{Problem Formulation}

In a vanilla GCN model (see the approximations \cite{kwSSC} based on \cite{dbCNN}),
the network architecture is recursively defined by
\begin{equation}\label{eq:gcn}
{H}^{l+1} = \sigma\left(\tilde{A} {H}^l {W}^l\right),\quad{}H^0=X,\nonumber
\end{equation}
where ${H}^{l}_{n\times{d^l}}$ is the feature matrix of the $l$'th layer
with its rows corresponding to the samples,
$W^l_{d^l\times{d}^{l+1}}$ is the sample-wise feature transformation matrix,
$\tilde{A}_{n\times{n}}$ is the normalized adjacency matrix so that
$\tilde{A} = (D+I)^{-\frac{1}{2}} (A+I) (D+I)^{-\frac{1}{2}}$,
$I$ is the identity matrix,
$D=\diag(A{1})$ is the degree matrix, $1$ is the vector of all ones,
$\diag()$ means a diagonal matrix \wrt the given diagonal entries,
and $\sigma$ is an element-wise nonlinear activation function.
Based on a given set of samples $X$ and optionally the corresponding labels $Y$,
learning is implemented by $\min_{W} \ell\left(X,Y,A,W\right)$,
where $\ell$ is a loss (\eg cross-entropy), usually expressed in terms of
$Y$ and $H^L$, the feature matrix obtained by stacking multiple GCN layers.

Our basic assumption is that $A$ is observed \wrt an underlying generative model
as well as some random observation noise.
In order to make learning robust to these noise and generalize well,
we minimize the expected loss
\begin{equation}\label{eq:bayesopt}
\min_{W} \int q(\phi\,\vert\,\varphi) \ell\left(X,Y,A(\phi),W\right) d\phi,
\end{equation}
where
$A(\phi)$ is a parameterization of graph adjacency matrices,
$A(0)=A$ is the original adjacency matrix,
$q(\phi\,\vert\,\varphi)$ is a zero-centered random perturbation
so that $A(\phi)$ in a ``neighborhood'' of $A$,
and $\varphi$ is the freedom of this perturbation.


To implement this machinery,
we must answer a set of fundamental questions: \ding{192} How to define
the manifold $\mathcal{M}$ of graphs, \ie the space of $A$?
\ding{193}~How to properly define the neighborhood $\{A(\phi):\phi\sim{}q(\phi\,\vert\,\varphi)\}$?
\ding{194}~What is the guiding principles to learn the neighborhood parameters $\varphi$?
We will build a geometric solution to these problems,
provide an efficient implementation of~\cref{eq:bayesopt},
and test the empirical improvement in generalization.
Our contributions are both theoretical and practical, which are summarized as follows:
\begin{itemize}
\item We bridge quantum information theory with graph neural networks,
and provide Riemannian metrics in closed form on the manifold of graphs;
\item We build a modified GCN~\cite{kwSSC} called the FisherGCN that can consistently
improve generalization;
\item We introduce \cref{alg:highorder} to pre-process the graph adjacency matrix for GCN
so as to incorporate high order proximities.
\end{itemize}

The rest of this paper is organized as follows.
We first review related works in \cref{sec:related}.
\Cref{sec:geo} introduces basic quantum information geometry,
based on which the following \cref{sec:intrinsic} formulates the manifold of graphs $\mathcal{M}$ and graph neighborhoods.
\Cref{sec:fishergcn,sec:exp} present the technical details and experimental results of our proposed FisherGCN.
\Cref{sec:extrinsic,sec:embed} provide our theoretical analysis on different ways to define the geometry of $\mathcal{M}$.
\Cref{sec:con} concludes and discusses future extensions.

\section{RELATED WORKS}\label{sec:related}

Below, we related our work to deep learning on graphs (with a focus on sampling strategies),
adversary learning, and quantum information geometry.

\subsection*{Graph Neural Networks}

The graph convolutional network~\cite{kwSSC} is a state-of-the-art graph neural network~\cite{gmANM,sgTGN} which performs convolution on the graphs in the spectral domain. While the performance of GCNs is very attractive, spectral convolution is a costly operation. Thus, the most recent implementations, \eg GraphSAGE~\cite{hyIRL}, takes convolution from spectral to spatial domain defined by the local neighborhood of each node. The average pooling on the nearest neighbors of each node is performed to capture the contents of the neighborhood. Below we describe related works which, one way or another, focus on various sampling strategies to improve aggregation and performance.

\subsection*{Structural Similarity and Sampling Strategies}

Graph embeddings~\cite{paDOL,glNSF} capture structural similarities in the graph. DeepWalk~\cite{paDOL} takes advantage of simulated localized walks in the node proximity which are then forwarded to the language modeling neural network to form the node context.
Node2Vec~\cite{glNSF}  interpolates between breadth- and depth-first sampling strategies to aggregate different types of neighborhood.

MoNet~\cite{mbGDL} generalizes the notion of coordinate spaces by learning a set of parameters of Gaussian functions to encode some distance for the node embedding, \eg the difference between degrees of a pair of nodes. Graph attention networks~\cite{vcGAN} learn such weights via a self-attention mechanism.
Jumping Knowledge Networks (JK-Nets) \cite{pmlr-v80-xu18c} also target the notion of node locality. Experiments on JK-Nets show that depending on the graph topology, the notion of the subgraph neighborhood varies, \eg random walks progress at different rates in different graphs. Thus, JK-Nets aggregate over various neighborhoods 
and considers multiple node localities.
By contrast, we apply mild adversary perturbations of the graph Laplacian based on quantum Fisher information so as to improve generalization. Thus, we infer a ``correction'' of the Laplacian matrix while JK-Net aggregates multiple node localities.


Sampling strategy has also an impact on the total size of receptive fields. In the vanilla GCN~\cite{kwSSC},  the receptive field of a single node grows exponentially \wrt
the number of layers which is computationally costly and results in so-called over smoothing of signals~\cite{lhDII}. 
Thus, stochastic GCN \cite{czSTG} controls the variance of the activation estimator by keeping the history/summary of activations in the previous layer to be reused.

Both our work and Deep Graph Infomax (DGI)~\cite{velickovic2018deep} take an information theoretic approach.
DGI maximizes the mutual information between representations of local subgraphs (a.k.a. patches) and high-level summaries of graphs while minimizing the mutual information between negative samples and the summaries. This ``contrasting'' strategy is somewhat related to our approach as we generate adversarial perturbation of the graph to flatten the most abrupt curvature directions. DGI relies on the notion of positive and negative samples. In contrast, we learn maximally perturbed parameters of our extrinsic graph representation which are the analogy to negative samples.

Lastly, noteworthy are application driven pipelines, \eg for molecule classification~\cite{dmCNG}.

\subsection*{Adversarial Learning}

The role of adversarial learning is to generate difficult-to-classify data samples by identifying them along the decision boundary and ``pushing'' them over this boundary. In a recent DeepFool approach~\cite{deepFool}, a cumulative sparse adversarial pattern is learned to maximally confuse predictions on the training dataset. Such an adversarial pattern generalizes well to confuse prediction on test data. Adversarial learning is directly connected to sampling strategies, \eg sampling hard negatives (obtaining the most difficult samples), and it has been long investigated~\cite{gpGAN,sLFC} in the community, especially in the shallow setting \cite{svm_adv1,attr_adver,svm_adv2}.

Adversarial attacks under the Fisher information metric (FIM)~\cite{zfTAA} propose to carry out perturbations in the spectral domain. Given a quadratic form of the FIM, the optimal adversarial perturbation is given by the first eigenvector corresponding to the largest eigenvalue. The larger the eigenvalues of the FIM are, the larger is the susceptibility of the classification approach to attacks on the corresponding eigenvectors.

Our work is related in that we also construct a quantum version of the FIM \wrt a parameterization of the graph Laplacian. We perform a maximization \wrt these parameters to condition the FIM around the local optimum, thus making our approach well regularized in the sense of flattening the most curved directions associated with the FIM. With the smoothness constraint, the classification performance typically degrades, \eg see the impact of smoothness on kernel representations~\cite{ckns}. Indeed, study \cite{robustness_odds} further shows there is a fundamental trade-off between high accuracy and the adversarial robustness.

However, our min-max formulation seeks the most effective perturbations (according to  \cite{zfTAA}) which thus simultaneously prevents unnecessary degradation of the decision boundary. With robust regularization for medium size datasets, we avoid overfitting which boosts our classification performance, as demonstrated in the following \cref{sec:exp}.

\subsection*{Quantum Information Geometry}

Natural gradient~\cite{aIGI,aFIA,pbRNG,zsNNG,snRFI} is a second-order optimization
procedure which takes the steepest descent \wrt the Riemannian geometry defined
by the FIM, which takes small steps on the directions with a large scale of FIM.
This is also suggestive that the largest eigenvectors of the FIM
are the most susceptible to attacks.




Bethe Hessian~\cite{skSCO}, or deformed Laplacian,
was shown to improve the performance of spectral clustering
on a par with non-symmetric and higher dimensional operators,
yet, drawing advantages of symmetric positive-definite representation.
Our graph Laplacian parameterization also draws on this view.

Tools from quantum information geometry are applied to machine learning~\cite{bjOTB,mcGPE}
but not yet ported to the domain of graph neural networks.
In information geometry, one can have different matrix divergences~\cite{nbMIG} that can be applied on the cone of p.s.d. matrices.
We point the reader to related definitions of the discrete Fisher information \cite{clADS} without illuminating the details.

\section{PREREQUISITES}\label{sec:geo}

\subsection*{Fisher Information Metric}

The discipline of information geometry~\cite{aIGI} studies the space of
probability distributions based on the Riemannian geometry framework.
As the most fundamental concept, the Fisher information matrix is
defined \wrt a given statistical model, \ie a parametric form of
the conditional probability distribution $p(X\,\vert\,\Theta)$, by
\begin{equation}\label{eq:fim}
\mathcal{G}(\Theta) =
\int p(X\,\vert\,\Theta)
\frac{\log p(X\,\vert\,\Theta)}{\partial\Theta}
\frac{\log p(X\,\vert\,\Theta)}{\partial\Theta^\top} \dx.
\end{equation}
By definition, we must have $\mathcal{G}(\Theta)\succeq0$.
Following H.~Hotelling and C.~R.~Rao,
this $\mathcal{G}(\Theta)$ is used (see section 3.5~\cite{aIGI} for history)
to define the Riemannian metric of a statistical model
$\mathcal{M}=\{\Theta\,:\,p(X\,\vert\,\Theta)\}$,
which is known as the Fisher information metric
$ds^2 = d\Theta^\top \fim(\Theta) d\Theta$.
Intuitively, the scale of $ds^2$ corresponds to the \emph{intrinsic} change of the model \wrt the movement $d\Theta$.
The FIM is invariant to reparameterization and is the unique Riemannian metric
in the space of probability distributions under certain conditions~\cite{cSDR,aIGI}.

\subsection*{Bures Metric}\label{sec:geobur}

In quantum mechanics, a quantum state is represented by
a graph (see \eg~\cite{bgsTL}). Denote a parametric graph Laplacian
matrix as $L(\Theta)$, and the trace-normalized Laplacian
$\rho(\Theta) = \frac{1}{\trace(L(\Theta))}L(\Theta)$
is known as the \emph{density matrix},
where $\trace(\cdot)$ means the trace.
One can therefore generalize the FIM to define a geometry of the $\Theta$ space.
In analogy to \cref{eq:fim},
the quantum version of the Fisher information matrix is
\begin{equation}\label{eq:qfim}
\fim_{ij}(\Theta)
=
\frac{1}{2} \trace\left[
\rho(\Theta)
(\partial{L}_i \partial{L}_j + \partial{L}_j \partial{L}_i)
\right],
\end{equation}
where $\partial{L}_i$ is the symmetric logarithmic derivative that
generalizes the notation of the derivative of logarithm:
\begin{equation*}
\frac{\partial\rho}{\partial\theta_i}
= \frac{1}{2} (\rho \cdot \partial{L}_i + \partial{L}_i \cdot \rho).
\end{equation*}
Let $\rho(\Theta)$ be diagonal, then
$\partial{L}_i={\partial \log\rho}/\partial\theta_i$.
Plugging into \cref{eq:qfim} will recover the traditional Fisher information defined in \cref{eq:fim}.
The quantum Fisher information metric $ds^2=d\Theta^\top\fim(\Theta)d\Theta$, up to constant scaling,
is known as the Bures metric~\cite{bAEO}. We use BM to denote these equivalent metrics and
abuse $\fim$ to denote both the BM and the FIM.
We develop upon the BM without considering its meanings in quantum mechanics.
This is because \ding{192} it can fall back to classical Fisher information;
\ding{193} its formulations are well-developed and can be useful
to develop deep learning on graphs.

\section{AN INTRINSIC GEOMETRY\label{sec:intrinsic}}

In this section, we define an intrinsic geometry of graphs based on the BM, so that one
can measure distances on the manifold of all graphs with a given number of nodes
and have the notion of \emph{neighborhood}.

We parameterize a graph by its density matrix
\begin{align}\label{eq:lcan}
\rho &= U \diag(\lambda) U^\top
= \sum_{i=1}^n \lambda_i u_i u_i^\top\succeq0,
\end{align}
where $UU^T=U^TU=I$ so that $U$ is on the unitary group,
\ie the manifold of unitary matrices, $u_i$ is
the $i$'th column of $U$,
$\lambda$ satisfies
$\lambda\ge0$, $\lambda^\top1=1$ and is on the closed probability simplex.
Notice that
the $\tau\ge1$ smallest eigenvalue(s) of the graph Laplacian
(and $\rho$ which shares the same spectrum up to scaling) are zero,
where $\tau$ is the number of connected components of the graph.

Fortunately for us, the BM \wrt this canonical parameterization
was already derived in closed form (see eq.(10)~\cite{hECO}), given by
\begin{equation}\label{eq:bm}
ds^2
= \frac{1}{2}\sum_{j=1}^n \sum_{k=1}^n \frac{(u_j^\top d\rho u_k)^2}{\lambda_j+\lambda_k}.
\end{equation}
For simplicity, we are mostly interested in the diagonal blocks of the FIM.
Plugging
\begin{equation*}
d\rho = \sum_{i=1}^n
\left[ d\lambda_i u_i u_i^\top
+  \lambda_i du_i u_i^\top
+ \lambda_i u_i du_i^\top
\right]
\end{equation*}
into \cref{eq:bm}, we get the following theorem.

\begin{theorem}\label{thm:bm}
In the canonical parameterization $\rho=U\diag(\lambda)U^\top$, the BM is
\begin{align*}
ds^2 &=
d\lambda^\top\fim(\lambda)d\lambda + \sum_{i=1}^n du_i^\top \fim(u_i) du_i\nonumber\\
&=
\sum_{i=1}^n \bigg[
\frac{1}{4\lambda_i} d\lambda_i^2 + d\lambda_i c_i^\top du_i\nonumber\\
&+
\frac{1}{2}
du_i^\top
\sum_{j=1}^n
\left(
\frac{(\lambda_i-\lambda_j)^2}{\lambda_i+\lambda_j}
u_j u_j^\top
\right)
du_i \bigg],
\end{align*}
where $c_i$ are some coefficients which we do not care about
that can be ignored in this paper.
\end{theorem}
One can easily verify that the first term in \cref{thm:bm}
coincides with the simplex geometry induced by the FIM.
Note that the BM is invariant to reparameterization, and we can write it in the following equivalent form.
\begin{corollary}\label{thm:bmtheta}
Under the reparameterization $\lambda_i=\exp(\theta_i)$
and $\rho(\theta,U)=U\diag(\exp(\theta))U^\top$, the BM is
\begin{align*}
ds^2 &= \sum_{i=1}^n \bigg[
\frac{\exp(\theta_i)}{4} d\theta_i^2
+ \exp(\theta_i)d\theta_i c_i^\top du_i\nonumber\\
&+ \frac{1}{2}
du_i^\top
\sum_{j=1}^n
\left[
\frac{(\exp(\theta_i)-\exp(\theta_j))^2}
{\exp(\theta_i)+\exp(\theta_j)}
u_j u_j^\top
\right]
du_i \bigg].
\end{align*}
\end{corollary}
This parameterization is favored in our implementation because after a small movement in
the $\theta$-coordinates, the density matrix is still p.s.d.

The BM allows us to study \emph{quantatively} the intrinsic change of the graph 
measured by $ds^2$.
For example, a constant scaling of the edge weights
results in $ds^2=0$ because the density matrix
does not vary.
The BM of the eigenvalue $\lambda_i$ is
proportional to $1/\lambda_i$, therefore as
the network scales up and $n\to\infty$,
the BM of the spectrum will scales up.
By the Cauchy-Schwarz inequality, we have
\begin{equation}
\trace(\fim(\lambda))
= \frac{1}{4}(1^\top\lambda^{-1}) (1^\top\lambda)
\ge{}\frac{n^2}{4}.
\end{equation}
It is, however, not straightforward to see
the scale of $\fim(u_i)$, that is
the BM \wrt the eigenvector $u_i$.
We therefore have the following result.
\begin{corollary}\label{thm:bmscale}
$\trace(\fim(u_i))
=
\frac{1}{2}
\sum_{j=1}^n \frac{(\lambda_i-\lambda_j)^2}%
{\lambda_i+\lambda_j}\le\frac{1}{2}$;\\
$\trace(\fim(U)) =
\frac{1}{2}
\sum_{i=1}^n
\sum_{j=1}^n \frac{(\lambda_i-\lambda_j)^2}%
{\lambda_i+\lambda_j}\le\frac{n}{2}$.
\end{corollary}
\begin{remark}
The scale (measured by trace) of $\fim(U)$ is $O(n)$
and $U$ has $O(n^2)$ parameters. The scale of $\fim(\lambda)$ is $O(n^2)$ and $\lambda$ has $(n-1)$ parameters. 
\end{remark}
Therefore, informally, the $\lambda$ parameters carry more information than $U$. Moreover,
it is computationally more expensive to parameterize $U$.
We will therefore make our perturbations on the spectrum $\lambda$.

We need to make a low-rank approximation of $\rho$ so as to reduce the degree of freedoms,
and make our perturbation cheap to compute. Based on the Frobenius norm, the best low-rank approximation
of any given matrix can be expressed by its largest singular values and their corresponding singular vectors.
Similar results hold for approximating density matrix based on the BM.
While BM is defined on an infinitesimal neighborhood, its corresponding non-local distance is known as
the Bures distance $D_B(\rho_1,\rho_2)$ given by
\begin{equation*}
D_B^2(\rho_1,\rho_2)=2\left(1-\trace\left(\sqrt{\rho_1^{\frac{1}{2}}\rho_2\rho_1^{\frac{1}{2}}}\right)\right).
\end{equation*}
For diagonal matrices, the Bures distance reduces to the Hellinger distance up to
constant scaling.

We have the following low-rank projection of a given density matrix.
\begin{theorem}\label{thm:lowerank}
Given $\rho_0=U\diag(\lambda)U^\top$, where $\lambda_1,\cdots\lambda_n$ are monotonically non-increasing,
its $\mathrm{rank}$-k projection is
\begin{equation*}
\bar{\rho}_0^k =
\argmin_{\rho:\mathrm{rank}(\rho)=k} D_B(\rho,\rho_0)
=
\frac{\sum_{i=1}^k\lambda_{i}u_iu_i^\top}{\sum_{i=1}^k\lambda_{i}}.
\end{equation*}
\end{theorem}
Our proof in the supplementary material\footnote{The supplementary material is in the appendix of \url{https://arxiv.org/abs/1903.04154}.
Our codes to reproduce all reported experimental results are available at \url{https://github.com/stellargraph/FisherGCN}.}
is based on Theorem 3~\cite{mmQSD}.  We may simply denote $\bar{\rho}_0^k$ as $\bar{\rho}_0$ with the spectrum decomposition
$\bar{\rho}_0=\bar{U}_0\diag(\bar{\lambda}_0)\bar{U}_0^\top$.


Hence, we can define a neighborhood of $A$ by varying the spectrum of
$\bar{\rho}^k(A)$. Formally, the graph Laplacian of the perturbed $A$ 
\wrt the perturbation $\phi$ is
\begin{equation}\label{eq:laphi}
L(A(\phi)) = \trace(L(A))\;\rho(A(\phi))
\end{equation}
so that its trace is not affected by the perturbation, and the perturbed density matrix is
\begin{equation}\label{eq:rhoaphi}
\rho(A(\phi)) =
\rho(A) + \bar{U}\diag\left(\frac{\exp(\bar{\theta}+\phi)}{1^\top\exp(\bar{\theta}+\phi)} -\bar{\lambda}\right)\bar{U}^\top,
\end{equation}
where the second low-rank term on the rhs is a perturbation of $\bar{\rho}^k(A)$
whose trace is 0 so that $\rho(A(\phi))$
is still a density matrix.
The random variable $\phi$ follows
\begin{equation}\label{eq:qphiisotropic}
q_{\mathrm{iso}}(\phi) = \mathcal{U}\left(\phi\,\vert\,0,\fim^{-1}(\bar{\theta})\right),
\end{equation}
which can be either a Gaussian distribution or a uniform distribution%
\footnote{Strictly speaking, $\mathcal{U}(\phi\,\vert\,\mu,\Sigma)$ should be the pushforward distribution \wrt the
Riemannian exponential map, which maps the distribution on the tangent space to the parameter manifold.},
which has zero mean and precision matrix $\fim(\bar{\theta})$ up to constant scaling.
Intuitively, it has smaller variance on the directions with a large $\fim$,
so that $q(\phi)$ is intrinsically isotropic \wrt the BM.

In summary, our neighborhood of a graph with adjacency matrix $A$
has $k$ most informative dimensions selected by the BM, and is defined by \cref{eq:laphi,eq:rhoaphi,eq:qphiisotropic}.
To compute this neighborhood, one needs to pre-compute the $k$ largest eigenvectors of $\rho(A)$,
which can be performed efficiently~\cite{mmRBK} for small $k$.
LanczosNet~\cite{lzLMS} also utilizes an eigendeompositiona sub-module for a low-rank approximation
of the graph Laplacian. Their focus is on building spectral filters rather than geometric perturbations.
An empirical range of $k$ is $10\sim50$.

One may alternatively parameterize a neighborhood by corrupting the graph links.
However, it is hard to control the scale of the perturbation based on information theory
and to have a compact parameterization.

\section{FISHER-BURES ADVERSARY GCN}\label{sec:fishergcn}

Based on the previous \cref{sec:intrinsic},
we know how to define the graph neighborhood.
Now we are ready to implement our perturbed GCN,
which we call the ``FisherGCN''.

We parameterize the perturbation as
\begin{equation}\label{eq:repara}
\phi(\varphi,\varepsilon)
= \fim^{-1/2}(\bar{\theta}) \diag(\varphi) \varepsilon
= \exp\left(-\frac{\bar{\theta}}{2}\right)\circ\varphi\circ\varepsilon,
\end{equation}
where ``$\circ$'' means element-wise product,
$\epsilon$ follows the uniform distribution over $[-\frac{1}{2},\frac{1}{2}]^k$
or the multivariate Gaussian distribution,
and \cref{thm:bmtheta} is used here to get $\fim(\bar{\theta})$.
The vector $0<\varphi\le\epsilon$ contains shape parameters
(One can implement the constraint through reparameterization $\varphi=\epsilon/(1+\exp(-\xi))$),
where $\epsilon$ is a hyper-parameter specifying the radius of the perturbation.
If $\varphi=\epsilon1$, then
$\phi$ follows $q_{\mathrm{iso}}(\phi)$ in \cref{eq:qphiisotropic}.
Then, one can compute the randomly perturbed density matrix $\rho(A(\phi))$ and corresponding
Laplacian matrix $L(A(\phi))$ based on \cref{eq:laphi,eq:rhoaphi}.

Our learning objective is to make predictions that is robust to such graph perturbations
by solving the following minimax problem
\begin{equation}\label{eq:l}
\min_{W} \max_{\varphi}
-\frac{1}{NM}\sum_{i=1}^N \sum_{j=1}^M \log p(Y_i\,\vert\,X_i, A(\phi(\varphi,\varepsilon_j)), W),
\end{equation}
where $M$ (\eg $M=5$) is the number of perturbations.
Similar to the training procedure of a GAN~\cite{gpGAN}, one can solve the optimization problem
by alternatingly updating $\varphi$ along $\bigtriangledown\varphi$, the gradient \wrt $\varphi$,
and updating $W$ along $-\bigtriangledown{W}$.

For brevity, we highlight the key equations and steps (instead of a full workflow)
of FisherGCN as follows:
\begin{enumerate}
\item[\ding{192}] Normalize $A$ (use the renormalization trick~\cite{kwSSC} or our \cref{alg:highorder} that will be introduced in \cref{sec:exp});
\item[\ding{193}] Compute $\bar{\rho}^k(A)$ (\cref{thm:lowerank}) by sparse matrix factorization
~\cite{mmRBK}
(only the top $k$ eigenvectors of $\rho(A)$ is needed, and this needs only to be done for once);
\item[\ding{194}] Perform regular GCN optimization
\begin{enumerate}
\item Use \cref{eq:repara} to get the perturbation $\phi$;
\item Use \cref{eq:laphi,eq:rhoaphi} to get the perturbed density matrix
$\rho(A(\phi))$ and the graph Laplacian matrix $L(A(\phi))$;
\item Plug $\tilde{A} = I-\trace(L(A)) \rho(A(\phi))$ into \cref{eq:gcn}.
\end{enumerate}
\end{enumerate}
Notice that the $A$ matrix (and the graph Laplacian) is normalized
in step \ding{192}
before computing the density matrix, so that the multiple
multiplications with $A$ in different layers do not cause numerical instability.
This can be varied depending on the implementation.

Our loss only imposes $k$ (\eg $k=10\sim50$) additional free parameters
(the rank of the projected $\bar{\rho}^k(A)$),
while $W$ contains the majority of the free parameters.
As compared to GCN, we need to solve the $k$ leading eigenvectors of
$\rho(A)$ before training, and multiply the computational cost
of training by a factor of $M$.
Notice that $\rho(A)$ is sparse and the eigendecomposition of sparse matrix
only need to be performed once.
Instead of computing the perturbed density matrix explicitly,
which is not sparse anymore,
one only need to compute the correction term
\begin{equation*}
\left[ \bar{U}_{n\times{k}}\diag\left(\frac{\exp(\bar{\theta}+\phi)}{1^\top\exp(\bar{\theta}+\phi)} -\bar{\lambda}\right)\bar{U}_{k\times{n}}^\top \right] X_{n\times{D}}
\end{equation*}
which can be solved efficiently in $O(knD)$ time.
If $k$ is small, this computational cost can be ignored (with no increase in the overall complexity)
as computing $AX$ has $O(md)$ complexity ($m$ is the number of links).
In summary, our FisherGCN is several times slower than GCN with roughly the same number of free parameters and complexity.

FisherGCN can be intuitively understood as running multiple GCN in parallel,
each based on a randomly perturbed graph. To implement the method does not
require understanding our geometric theory but only to follow
the list of pointers \ding{192}\ding{193}\ding{194} shown above.

\section{EXPERIMENTS}\label{sec:exp}

\begin{table*}[t]
\caption{Dataset statistics. Note the number of links reported in previous works~\cite{kwSSC} counts duplicate
links and some self-links, which is corrected here. ``\#Comps'' means the number of connected components.
``Sparsity'' shows the sparsity of the matrix $\tilde{A}$. ``Sparsity$^T$'' shows its sparsity in GCN$^T$ (with mentioned settings of $T$ and $\epsilon$).\label{tbl:data}}
\centering
\begin{tabular}{crrcrcrcc}
\hline Dataset & \#Nodes & \#Links & \#Comps & \#Features & \#Classes & Train:Valid:Test & Sparsity & Sparsity$^T$\\
\hline
Cora     &  2,708 & 5,278 & 78 & 1,433 & 7 & 140:500:1000
& 0.18\% & 9.96\% \\
CiteSeer &  3,327 & 4,552 & 438 & 3,703 & 6 & 120:500:1000
& 0.11\% & 3.01\% \\
PubMed   &  19,717 & 44,324 & 1 & 500 & 3 & 60:500:1000
& 0.03\% & 3.31\% \\
\hline
\end{tabular}
\vspace{1em}
\caption{Testing loss and accuracy in percentage.
The hyperparameters (learning rate 0.01; 64 hidden units; dropout rate 0.5; weight decay $5\times10^{-4}$)
are selected based on the best overall testing accuracy of GCN on Cora and CiteSeer.
Then we use these hyperparameters across all the four methods and three datasets.
The reported mean$\pm$std scores are based on 200 runs
(20 random splits; 10 different initializations per split).
The splits used for hyperparameter selection and testing are different.\label{tbl:results}}
\begin{tabular}{c|ccc|ccc}
\hline
& \multicolumn{3}{|c|}{Testing Accuracy} & \multicolumn{3}{|c}{Testing Loss} \\
\hline & Cora & CiteSeer & PubMed & Cora & CiteSeer & PubMed \\
\hline
GCN              & $80.52\pm2.3$ & $69.59\pm2.0$ & $78.17\pm2.4$
                 & $1.07\pm0.04$ & $1.36\pm0.03$ & $0.75\pm0.04$ \\
FisherGCN        & $80.70\pm2.2$ & $69.80\pm2.0$ & $78.43\pm2.4$
                 & $1.06\pm0.04$ & $1.35\pm0.03$ & $0.74\pm0.04$ \\
GCN$^T$          & $81.20\pm2.3$ & $70.31\pm1.9$ & $78.99\pm2.6$
                 & $1.04\pm0.04$ & $1.33\pm0.03$ & $0.70\pm0.05$ \\
FisherGCN$^T$    & $\bm{81.46}\pm2.2$ & $\bm{70.48}\pm1.7$ & $\bm{79.34}\pm2.7$
                 & $\bm{1.03}\pm0.03$ & $\bm{1.32}\pm0.03$ & $\bm{0.69}\pm0.04$ \\
\hline
\end{tabular}
\end{table*}

In this section, we perform an experimental study on
semi-supervised transductive node classification tasks.
We use three benchmark datasets, namely, the Cora, CiteSeer and PubMed citation networks~\cite{ycRSS,kwSSC}.
The statistics of these datasets are displayed in \cref{tbl:data}.
As suggested recently~\cite{smPOG}, we use random splits of training:validation:testing datasets
based on the same ratio as the Planetoid split~\cite{ycRSS}, as given in the ``Train:Valid:Test'' column in \cref{tbl:data}.

We will mainly compare against GCN which can represent the state-of-the-art on
these datasets, because our method serves as an ``add-on'' of GCN.
We will discuss how to adapt this add-on to other graph neural networks in \cref{sec:con}
and refer the reader to \cite{smPOG} for how the performance of GCN compares against the other methods.
Nevertheless, we introduce a stronger baseline called GCN$^T$.
It was known that random walk similarities can help improve
learning of graph neural networks~\cite{yhGCN}.
We found that pre-processing
the graph adjacency matrix $A$ (with detailed steps listed in \cref{alg:highorder})
can improve the performance of GCN on semisupervised
node classification tasks\footnote{During the review period of this paper,
related works appeared~\cite{wsSGC,apMHO} which build a high order GCN.
Comparatively, our GCN$^T$ is closely based on DeepWalk similarities~\cite{paDOL}
instead of power transformations of the adjacency matrix.}
This processing is based on DeepWalk similarities~\cite{paDOL}
that are explicitly formulated in Table~1~\cite{qdNEA}.
\Cref{alg:highorder} involves two hyperparameters:
the order $T\ge1$ determines the order of the proximities
(the larger, the denser the resulting $A$; $T=1$ falls back to the regular GCN);
the threshold $\nu>0$ helps remove links with small probabilities to enhance sparsity.
In the experiments we fix $T=5$ and $\nu=10^{-4}$.
These procedures correspond to a polynomial filter with hand-crafted coefficients.
One can look at \cref{tbl:data} and compare the sparsity of the processed adjacency matrix
by \cref{alg:highorder} (in the ``Sparsity$^T$'' column) v.s.
the original sparsity (in the ``Sparsity'' column) to have a rough idea
on the computational overhead of GCN$^T$ v.s. GCN.

Our proposed methods are denoted as FisherGCN and FisherGCN$^T$,
which are respectively based on GCN and GCN$^T$.
We fix the perturbation radius parameter $\epsilon=0.1$ and the rank parameter $k=10$.

\begin{algorithm}
$A \leftarrow \mathrm{diag}^{-1}(A1) A$\;
$S,B \leftarrow A$\;
\For{$t \gets 2$ \textbf{to} $T$}{
  $B\leftarrow  B A$\;
  $S\leftarrow S + B$\;
}
$A \leftarrow \frac{1}{T}S \circ (1_{n\times{n}}-I)$\;
$A \leftarrow A \circ (A>\nu)$\;
$A \leftarrow A + A^\top + 2I$\;
$A \leftarrow \mathrm{diag}^{-\frac{1}{2}}(A1)\;A\;\mathrm{diag}^{-\frac{1}{2}}(A1)$\;
\caption{Pre-process $A$ to capture high-order proximities ($T\ge2$ is the order; $\nu>0$ is a threshold)\label{alg:highorder}}
\label{algo:max}
\end{algorithm}

The testing accuracy and loss are reported in \cref{tbl:results}.
We adapt the GCN codes~\cite{kwSSC} so that the four methods are compared in exactly the same
settings and only differ in the matrix $A$ that is used for computing the graph convolution.
One can observe that FisherGCN and GCN$^T$
can both improve over GCN, which means that our perturbation
and the pre-processing by \cref{alg:highorder} both help to improve generalization.
The best results are given by FisherGCN$^T$ with both techniques added.
Th large variation is due to different splits of the training:validation:testing datasets~\cite{smPOG},
and therefore these scores vary with the split.
In repeated experiments, we observed a consistent improvement of the proposed methods as compared to the baselines.

\begin{figure}[t]
\centering
\includegraphics[width=.47\textwidth]{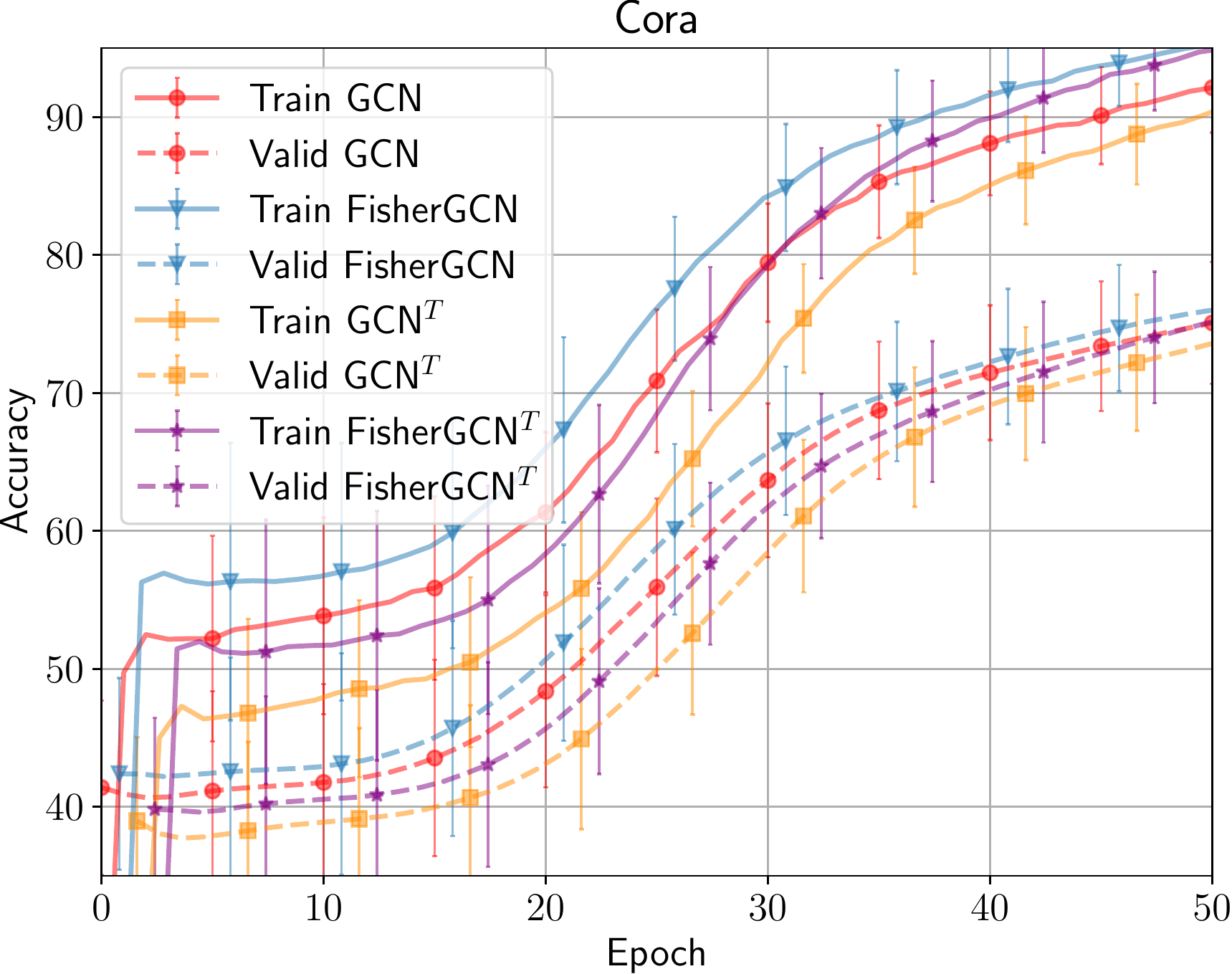}
\caption{Learning curves (averaged over 200 runs) in accuracy on the Cora dataset.\label{fig:cora}}
\end{figure}

\Cref{fig:cora} shows the learning curves on the Cora dataset (see the supplementary material for the other cases).
We can observe that the proposed perturbation presents higher training and testing scores during learning.
The performance boost of FisherGCN is more significant if the number of epochs is limited to a small value.

\section{AN EXTRINSIC GEOMETRY}\label{sec:extrinsic}

In this section and the following section \cref{sec:embed}, we present analytical results on
the geometry of the manifold of graphs. These results are useful to interpret the proposed FisherGCN
and are useful to understand graph-based machine learning.

We first derive an \emph{extrinsic} geometry of
a parametric graph embedded in a neural network.
Based on this geometry, the learner can capture the curved directions of the loss surface
and make more effective perturbations than the isotropic perturbation in \cref{eq:rhoaphi}.
While the intrinsic geometry in \cref{sec:intrinsic} measures
how much the graph itself has changed due to a movement on $\mathcal{M}$,
the extrinsic geometry measures how varying the parameters of the graph
will change the external model.
Intuitively, if a dynamic $\Delta{G}$ causes little change based on the
intrinsic geometry, one may also expect $\Delta{G}$ has little effect
on the external neural network. However, in general, these two geometries
impose different Riemannian metrics on the same manifold $\mathcal{M}$ of graphs.

Consider the predictive model represented by the conditional distribution
$p(Y\,\vert\,X,A(\phi),W)$. Wlog consider $\phi$ is a scalar,
which serves as a coordinate system of graphs.
We use $\fim^E$ to denote the extrinsic Riemannian metric
(the upper script ``$E$'' is for extrinsic)
that is to be distinguished with the intrinsic $\fim$.
Based on the GCN computation introduced in \cref{sec:intro},
we can get an explicit expression of $\fim^E$.
\begin{theorem}\label{thm:fim}
Let $\ell=-\log{p}(Y\,\vert\,X,A(\phi),W)$,
$\Delta_{l}={\partial\ell}/{\partial{}H^l}$
denote the back-propagated error of layer $l$'s output $H^l$,
and $\Sigma_{l}'$ denote the derivative of
layer $l$'s activation function. Then
\begin{align*}
&\fim^E(\phi)\nonumber\\
&=
\frac{1}{N}\sum_{i=1}^N
\left(
\sum_{l=0}^{L-1} \left(
H^lW^l
(\Delta_{l+1}\circ\Sigma_{l+1}')^\top
\frac{\partial\tilde{A}}{\partial\phi} \right)_{ii} \right)^2\nonumber\\
&=
\frac{1}{N}\sum_{i=1}^N \left(
\sum_{l=0}^{L-1} \left(
H^l \Delta_l^\top \frac{\partial\tilde{A}}{\partial\phi}
\right)_{ii} \right)^2;\\
&\fim^E(W^l)\nonumber\\
&=
\frac{1}{N}\sum_{i=1}^N \bigg(
\mathrm{vec}\left(
(\Delta_{l+1}\circ\Sigma_{l+1}')^\top\tilde{A} H^l
\right)\\
&\hspace{4em}\times\mathrm{vec}^\top\left(
(\Delta_{l+1}\circ\Sigma_{l+1}')^\top\tilde{A} H^l
\right) \bigg),
\end{align*}
where $\mathrm{vec}()$ means rearranging a matrix into a column vector.
\end{theorem}

The information geometry of neural networks is mostly used
to develop the second order optimization \cite{pbRNG,aFIA},
where $\fim^E(W)$ is used. Here we are mostly
interested in $\fim^E(\phi)$, and our target is not
for better optimization
but to find a neighborhood of a given graph with large intrinsic variations.
A movement with a large scale of $\fim^E$ can most effectively
change the predictive model $p(Y\,|\,X)$.

Let us develop some intuitions based on the term inside the trace
on the rhs of $\fim^E(\phi)$.
In order to change the predictive model,
the most effective edge increment $d{a}_{ij}$
should be positively correlated with $(h_i^{l\top} \Delta_{lj} + \Delta_{li}^{\top} h^l_j)$,
which means how the hidden feature of node $i$ (node $j$)
is correlated with the increment of the hidden feature of node $j$ (node $i$).
This makes intuitive sense.

\begin{figure}[!h]
\centering
\includegraphics[width=.2\textwidth]{graph.1}%
\includegraphics[width=.28\textwidth]{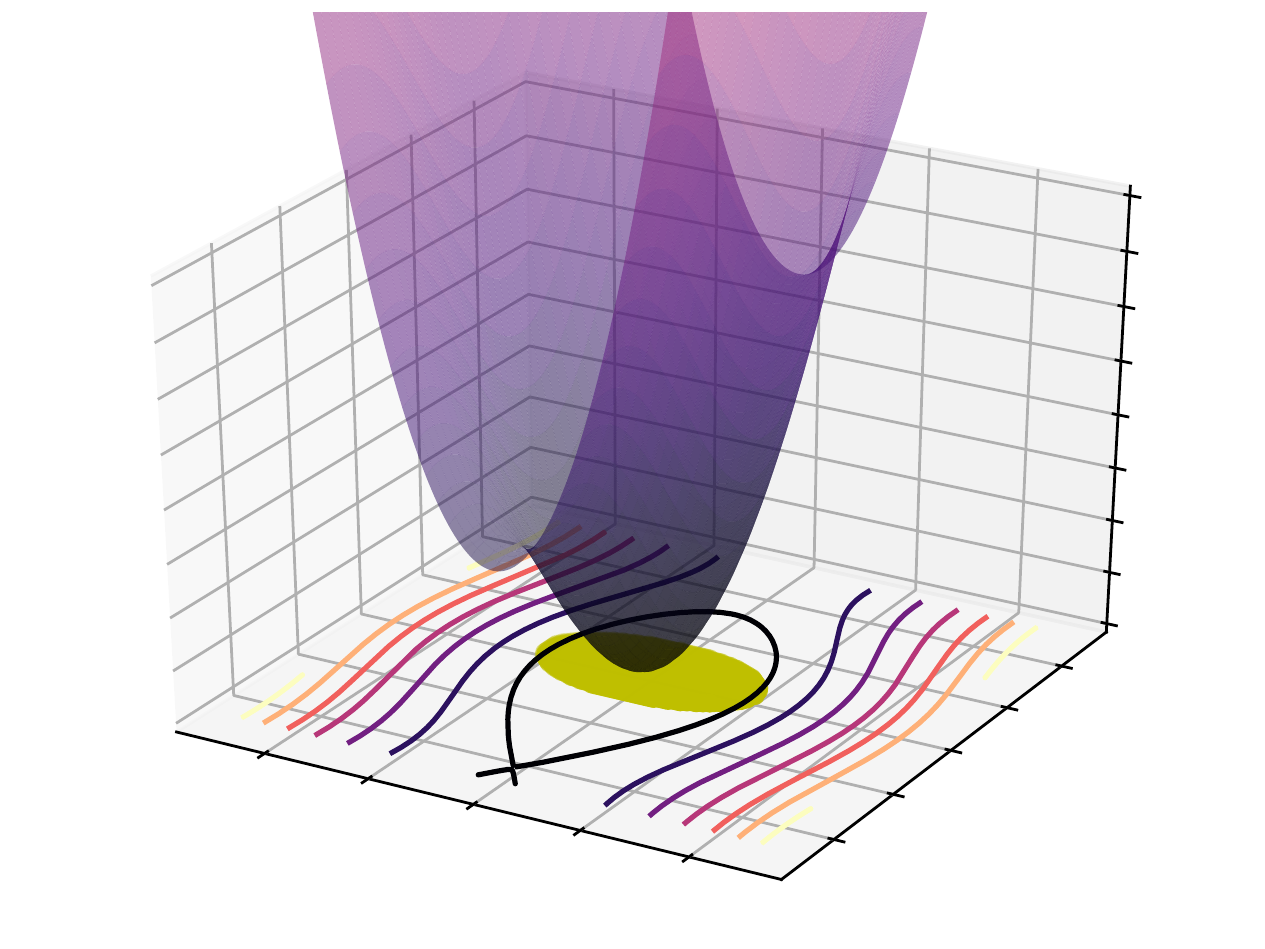}
\caption{\label{fig:neighbour}Learning a neighborhood (yellow region)
of a graph where the loss surface is most curved corresponding to large $\fim^E$.}
\end{figure}

The meaning of \cref{thm:fim} is mainly theoretical,
giving an explicit expression of $\fim^E$ for the GCN model,
which, to the best of the authors' knowledge, was not derived before
(most literature studies the FIM of a feed-forward model such as a multi-layer perceptron).
This could be useful for future works for natural gradient
optimizers specifically tailored for GCN. On the practical side,

\Cref{thm:fim} also helps to understand the proposed minimax optimization.
On the manifold $\mathcal{M}$ of graphs,
we make the rough assumption that $A(0)=A$ is a local minimum
of $\ell$ along the $\phi$ coordinate system, that is, adding
a small noise to $A$ will always cause an increment in the loss.
The random perturbation in \cref{eq:repara} corresponds to the distribution
\begin{equation*}
q(\phi\,\vert\,\varphi) = \mathcal{U}\left(\phi\,\vert\,0,\fim^{-1/2}(\bar{\theta}) \diag(\varphi\circ\varphi) \fim^{-1/2}(\bar{\theta})
\right),
\end{equation*}
and our loss function in \cref{eq:l} is obtained by applying the reparameterization trick~\cite{kwAEV}
to solve the expectation in \cref{eq:bayesopt}.
If $\varphi=\epsilon1$, then $q(\phi\,\vert\,\varphi)=\mathcal{U}(\phi\,\vert\,0, \epsilon^2\fim(\bar{\theta})^{-1})$
falls back to the isotropic $q_{\mathrm{iso}}(\phi)$.
Letting $\varphi$ free allows the neighborhood to deform (see
\cref{fig:neighbour} left).
Then, through the maximization in \cref{eq:l} \wrt $\varphi$,
the density $q(\phi)$ will focus on the neighborhood
of the original graph $A$ where the loss surface is most upcurved
(see \cref{fig:neighbour} right).
These directions have large $\fim^E$ and make the perturbation effective
in terms of the FIM of the graph neural network.
Consider the reverse case, when the density $q(\phi)$ corresponds to small values of $\fim^E$.
Such perturbations are long the flat directions of the loss surface and will
have little effect on learning the predictive model.

\section{AN EMBEDDING GEOMETRY}\label{sec:embed}

We present a geometry of graphs which is constructed in the spatial domain and
is closely related to graph embeddings~\cite{paDOL}.
Consider representing a graph by a node similarity matrix $W_{n\times{n}}$
(\eg based on \cref{alg:highorder}), which is row-normalized and has zero-diagonal entries.
These similarities are assumed to be based on a latent graph embedding $Y_{n\times{d}}$:
$p_{ij}(Y) = \frac{1}{Z_i} \exp\left( -\Vert y_i - y_j \Vert^2 \right)$,
where $P_{n\times{n}}$ is the generative model with the same constraints
as the $W$ matrix, and $Z_i$ is the partition function.
Then, the observed FIM (that leads to the FIM as the number of observations increase)
is given by the Hessian matrix of $\kl(W:P(Y))$ evaluated at the maximum
likelihood estimation $Y^\star=\argmin_{Y}\kl(W:P(Y))$,
where $\kl$ denotes the Kullback-Leibler divergence.
We have the following result.
\begin{theorem}\label{thm:embed}
W.r.t. the generative model $p_{ij}(Y)$,
the diagonal blocks of the observed FIM $\hat{\fim}$
of a graph represented by the similarity matrix $W$ is
\begin{align*}
\hat{\fim}(y^k)
=
& 4 L(W-P(Y)) + 8L({P}(Y)\circ{D}^k)\\
& - 4({B}^k)^\top {B}^k,
\end{align*}
where $y^k$ is the $k$'th column of $Y$,
$L(W-P(Y))$ is the Laplacian matrix computed based on the indefinite weights
$(W-P(Y))$ after symmetrization, $D^k=(y_{ik}-y_{jk})^2$, and $B^k=L(p_{ij}(y_{ik}-y_{jk}))$.
\end{theorem}
The theorem gives the observed FIM, while the expected FIM
(the 2nd and 3rd terms in \cref{thm:embed})
can be alternatively derived based on~\cite{smAIG}.
To understand this result, we can assume that $P(Y)\to{}W$ as the number of observations increase.
Then
\begin{small}
\begin{align*}
dy^{k\top} \hat{\fim}(y^k) dy^k =
4
\sum_{i=1}^n \bigg[\sum_{j=1}^n p_{ij}(y_{ik}-y_{jk})^2(dy_{ik}-dy_{jk})^2\nonumber\\
-
\left(\sum_{j=1}^n p_{ij}(y_{ik}-y_{jk})(dy_{ik}-dy_{jk})\right)^2
\bigg]
\end{align*}
\end{small}
is in the form of a variance of $(y_{ik}-y_{jk})(dy_{ik}-dy_{jk})=\frac{1}{2}d(y_{ik}-y_{jk})^2$
\wrt $p_{ij}$. Therefore a large Riemannian metric
$dy^{k\top} \hat{\fim}(y^k) dy^k$ corresponds to a motion $dy^k$
which cause a large variance of neighbor's distance increments.
For example, a rigid motion, or a uniform expansion/shrinking
of the latent network embedding will cause little or no effect
on the variance of $d(y_{ik}-y_{jk})^2$, and hence corresponds to
a small distance in this geometry.

This metric can be useful for developing theoretical perspectives of
network embeddings, or build spatial perturbations of
graphs (instead of our proposed spectral perturbation).
As compared to the intrinsic geometry in \cref{sec:intrinsic},
the embedding geometry is based on a generative model instead of the BM.
As compared to the extrinsic geometry in \cref{sec:extrinsic},
the embedding geometry is not related to a neural network model.

\section{CONCLUSION AND DISCUSSIONS}\label{sec:con}

We imported new tools and adapted the notations from quantum information geometry
to the area of geometric deep learning.
We discussed three different geometries on the ambient space of graphs,
with their Riemannian metrics provided in closed form.
The results and adaptations are useful to develop new deep learning methods.
We demonstrated their usage by perturbing graph structures in a GCN,
showing consistent improvements in transductive node classification tasks.

It is possible to generalize FisherGCN to a scalable setting, where a
mini-batch only contains a sub-graph~\cite{hyIRL} of $m\ll{n}$ nodes.
This is because our perturbation has a low-rank factorization
given by the second term in \cref{eq:rhoaphi}.
One can reuse this spectrum factorization of the global matrix
to build sub-graph perturbations.

If $A$ contains free-parameters~\cite{vcGAN}, one can compute
the low-rank projection $\bar{\rho}^k(A)$ using the
original graph that is parameter free, based on which
the perturbation term can be constructed.
Alternatively, one can periodically save the graph and recompute $\bar{\rho}^k(A)$ during learning.

Based on \cite{kwSSC}, we express a graph convolution operation on an input signal $x=\sum_{i=1}^n \alpha_i u_i\in\Re^n$ as
\begin{align*}
\left[ I - \trace(L) \rho(A) \right] x
&= x - \trace(L) E_{i\sim\lambda} (\alpha_i u_i),
\end{align*}
where $E$ denotes the expectation.
The von Neumann entropy
of the quantum state $\rho$ is defined by the Shannon entropy of
$\lambda$, that is $-\sum_{i=1}^n\lambda_i \log \lambda_i$.
If we consider a higher order convolutional operator
(in plain polynomial), given by
\begin{equation*}
\rho^\omega(A) x
=
\frac{1}{ \lambda^\omega 1 }
U\diag(\lambda)^\omega U^\top x
=
E_{i\sim\frac{\lambda^\omega}{\lambda^\omega1}} (\alpha_i u_i).
\end{equation*}
The von Neumann entropy is monotonically decreasing
as $\omega\ge1$ increases. As $\omega\to\infty$,
we have $\rho^\omega(A) x \to \alpha_1 u_1$
(if $\lambda_1$ is the largest eigenvalue of $\rho(A)$ without multiplicity).
Therefore, high order convolutions enhance the signal \wrt the largest eigenvectors of $\rho$.
Therefore our perturbation is equivalent to adding high order polynomial filters.
It is interesting to explore alternative perturbations based on other distances,
\eg matrix Bregman divergence~\cite{nmMMD}.  An empirical stay on comparing
different types of perturbations in the GCN setting is left as future work.

\subsubsection*{Acknowledgements}

The authors gratefully thank the anonymous UAI reviewers and Frank Nielsen
for their valuable and constructive comments.
We thank the authors~\cite{ycRSS,kwSSC,smPOG} for making their processed datasets and codes public.
This work is supported by StellarGraph@Data61.

\subsubsection*{References}
\renewcommand{\refname}{}
\vspace{-2.5em}
\bibliography{main}

\newcommand{\etalchar}[1]{$^{#1}$}
\begin{thebibliography}{AEHPK{\etalchar{+}}19}

\bibitem[AEHPK{\etalchar{+}}19]{apMHO}
S.~Abu-El-Haija, B.~Perozzi, A.~Kapoor, N.~Alipourfard, K.~Lerman,
  H.~Harutyunyan, G.~V. Steeg, and A.~Galstyan.
\newblock {M}ix{H}op: Higher-order graph convolutional architectures via
  sparsified neighborhood mixing.
\newblock In {\em ICML}, volume~97, pages 21--29. PMLR, 2019.

\bibitem[AKO19]{aFIA}
S.~Amari, R.~Karakida, and M.~Oizumi.
\newblock Fisher information and natural gradient learning in random deep
  networks.
\newblock In {\em AISTATS}, volume~89 of {\em PMLR}, pages 694--702, 2019.

\bibitem[Ama16]{aIGI}
S.~Amari.
\newblock {\em Information Geometry and Its Applications}, volume 194 of {\em
  Applied Mathematical Sciences}.
\newblock Springer-Verlag, Berlin, 2016.

\bibitem[BGS16]{bgsTL}
S.~L. Braunstein, S.~Ghosh, and S.~Severini.
\newblock The {L}aplacian of a graph as a density matrix: a basic combinatorial
  approach to separability of mixed states.
\newblock {\em Annals of Combinatorics}, 10(3):291--317, 2016.

\bibitem[BJL18]{bjOTB}
R.~Bhatia, T.~Jain, and Y.~Lim.
\newblock On the {B}ures{\textendash}{W}asserstein distance between positive
  definite matrices.
\newblock {\em Expositiones Mathematicae}, 2018.

\bibitem[BNL11]{svm_adv1}
B.~Biggio, B.~Nelson, and P.~Laskov.
\newblock Support vector machines under adversarial label noise.
\newblock In {\em ACML}, volume~20 of {\em PMLR}, pages 97--112, 2011.

\bibitem[Bur69]{bAEO}
D.~Bures.
\newblock An extension of {K}akutani's theorem on infinite product measures to
  the tensor product of semifinite $w^*$-algebras.
\newblock {\em Transactions of the AMS}, 135:199--212, 1969.

\bibitem[BZSL14]{bzSNA}
J.~Bruna, W.~Zaremba, A.~Szlam, and Y.~LeCun.
\newblock Spectral networks and locally connected networks on graphs.
\newblock In {\em ICLR}, 2014.

\bibitem[{\v{C}}en82]{cSDR}
N.~N. {\v{C}}encov.
\newblock {\em Statistical Decision Rules and Optimal Inference}, volume~53 of
  {\em Translations of Mathematical Monographs}.
\newblock American Mathematical Society, 1982.
\newblock (Published in Russian in 1972).

\bibitem[CLZ19]{clADS}
S.~Chow, W.~Li, and H.~Zhou.
\newblock A discrete {S}chr\"odinger bridge problem via optimal transport on
  graphs.
\newblock {\em Journal of Functional Analysis}, 276(8):2440--2469, 2019.

\bibitem[CZS18]{czSTG}
J.~Chen, J.~Zhu, and L.~Song.
\newblock Stochastic training of graph convolutional networks with variance
  reduction.
\newblock In {\em ICML}, volume~80 of {\em PMLR}, pages 942--950, 2018.

\bibitem[DBV16]{dbCNN}
M.~Defferrard, X.~Bresson, and P.~Vandergheynst.
\newblock Convolutional neural networks on graphs with fast localized spectral
  filtering.
\newblock In {\em NIPS}, pages 3844--3852. Curran Associates, Inc., 2016.

\bibitem[DMI{\etalchar{+}}15]{dmCNG}
D.~K. Duvenaud, D.~Maclaurin, J.~Iparraguirre, R.~Bombarell, T.~Hirzel,
  A.~Aspuru-Guzik, and R.~P Adams.
\newblock Convolutional networks on graphs for learning molecular fingerprints.
\newblock In {\em NIPS 28}, pages 2224--2232. Curran Associates, Inc., 2015.

\bibitem[GL16]{glNSF}
A.~Grover and J.~Leskovec.
\newblock {Node2Vec}: Scalable feature learning for networks.
\newblock In {\em KDD}, pages 855--864, 2016.

\bibitem[GMS05]{gmANM}
M.~Gori, G.~Monfardini, and F.~Scarselli.
\newblock A new model for learning in graph domains.
\newblock In {\em IJCNN}, volume~2, pages 729--734, 2005.

\bibitem[GPAM{\etalchar{+}}14]{gpGAN}
I.~Goodfellow, J.~Pouget-Abadie, M.~Mirza, B.~Xu, D.~Warde-Farley, S.~Ozair,
  A.~Courville, and Y.~Bengio.
\newblock Generative adversarial nets.
\newblock In {\em NIPS 27}, pages 2672--2680. Curran Associates, Inc., 2014.

\bibitem[H\"92]{hECO}
M.~H\"ubner.
\newblock Explicit computation of the {B}ures distance for density matrices.
\newblock {\em Physics Letters A}, 163(4):239 -- 242, 1992.

\bibitem[HYL17]{hyIRL}
W.~Hamilton, Z.~Ying, and J.~Leskovec.
\newblock Inductive representation learning on large graphs.
\newblock In {\em NIPS}, pages 1024--1034. Curran Associates, Inc., 2017.

\bibitem[KW14]{kwAEV}
D.~P. Kingma and M.~Welling.
\newblock Auto-encoding variational {B}ayes.
\newblock In {\em ICLR}, 2014.

\bibitem[KW17]{kwSSC}
T.~N. Kipf and M.~Welling.
\newblock Semi-supervised classification with graph convolutional networks.
\newblock In {\em ICLR}, 2017.

\bibitem[KXC11]{attr_adver}
M.~Kantarc{\i}o{\u{g}}lu, B.~Xi, and C.~Clifton.
\newblock Classifier evaluation and attribute selection against active
  adversaries.
\newblock {\em Data Mining and Knowledge Discovery}, 22(1):291--335, 2011.

\bibitem[LHW18]{lhDII}
Q.~Li, Z.~Han, and X.-M. Wu.
\newblock Deeper insights into graph convolutional networks for semi-supervised
  learning.
\newblock In {\em AAAI}, 2018.

\bibitem[LZUZ19]{lzLMS}
R.~Liao, Z.~Zhao, R.~Urtasun, and R.~S. Zemel.
\newblock {LanczosNet}: Multi-scale deep graph convolutional networks.
\newblock In {\em ICLR}, 2019.

\bibitem[MBM{\etalchar{+}}17]{mbGDL}
F.~Monti, D.~Boscaini, J.~Masci, E.~Rodol{\`{a}}, J.~Svoboda, and M.~M.
  Bronstein.
\newblock Geometric deep learning on graphs and manifolds using mixture model
  {CNN}s.
\newblock In {\em CVPR}, pages 5425--5434, 2017.

\bibitem[MC18]{mcGPE}
B.~Muzellec and M.~Cuturi.
\newblock Generalizing point embeddings using the {W}asserstein space of
  elliptical distributions.
\newblock In {\em NeurIPS 31}, pages 10237--10248. Curran Associates, Inc.,
  2018.

\bibitem[MFF16]{deepFool}
S.~Moosavi{-}Dezfooli, A.~Fawzi, and P.~Frossard.
\newblock {DeepFool}: A simple and accurate method to fool deep neural
  networks.
\newblock In {\em CVPR}, pages 2574--2582, 2016.

\bibitem[MKHS14]{ckns}
J.~Mairal, P.~Koniusz, Z.~Harchaoui, and C.~Schmid.
\newblock Convolutional kernel networks.
\newblock In {\em NIPS}, pages 2627--2635. Curran Associates, Inc., 2014.

\bibitem[MM15]{mmRBK}
C.~Musco and C.~Musco.
\newblock Randomized block {K}rylov methods for stronger and faster approximate
  singular value decomposition.
\newblock In {\em NIPS 28}, pages 1396--1404. Curran Associates, Inc., 2015.

\bibitem[MMPidZ08]{mmQSD}
D.~Markham, J.~Adam Miszczak, Z.~Pucha\l{}a, and K.~\ifmmode~\dot{Z}\else
  \.{Z}\fi{}yczkowski.
\newblock Quantum state discrimination: a geometric approach.
\newblock {\em Phys. Rev. A}, 77:042111, 2008.

\bibitem[NB13]{nbMIG}
F.~Nielsen and R.~Bhatia.
\newblock {\em Matrix Information Geometry}.
\newblock Springer-Verlag Berlin Heidelberg, 2013.

\bibitem[NMBN13]{nmMMD}
R.~Nock, B.~Magdalou, E.~Briys, and F.~Nielsen.
\newblock Mining matrix data with {B}regman matrix divergences for portfolio
  selection.
\newblock In F.~Nielsen and R.~Bhatia, editors, {\em Matrix Information
  Geometry}, pages 373--402. Springer Berlin Heidelberg, 2013.

\bibitem[PARS14]{paDOL}
B.~Perozzi, R.~Al-Rfou, and S.~Skiena.
\newblock {DeepWalk}: Online learning of social representations.
\newblock In {\em KDD}, pages 701--710, 2014.

\bibitem[PB14]{pbRNG}
R.~Pascanu and Y.~Bengio.
\newblock Revisiting natural gradient for deep networks.
\newblock In {\em ICLR}, 2014.

\bibitem[QDM{\etalchar{+}}18]{qdNEA}
J.~Qiu, Y.~Dong, H.~Ma, J.~Li, K.~Wang, and J.~Tang.
\newblock Network embedding as matrix factorization: Unifying {DeepWalk},
  {LINE}, {PTE}, and {Node2Vec}.
\newblock In {\em WSDM}, pages 459--467, 2018.

\bibitem[Sch92]{sLFC}
J.~Schmidhuber.
\newblock Learning factorial codes by predictability minimization.
\newblock {\em Neural Computation}, 4(6):863--879, 1992.

\bibitem[SGT{\etalchar{+}}09]{sgTGN}
F.~Scarselli, M.~Gori, A.~C. Tsoi, M.~Hagenbuchner, and G.~Monfardini.
\newblock The graph neural network model.
\newblock {\em IEEE Transactions on Neural Networks}, 20(1):61--80, 2009.

\bibitem[SKZ14]{skSCO}
A.~Saade, F.~Krzakala, and L.~Zdeborov\'{a}.
\newblock Spectral clustering of graphs with the {B}ethe {H}essian.
\newblock In {\em NIPS 27}, pages 406--414. Curran Associates, Inc., 2014.

\bibitem[SMBG18]{smPOG}
O.~Shchur, M.~Mumme, A.~Bojchevski, and S.~G\"unnemann.
\newblock Pitfalls of graph neural network evaluation.
\newblock In {\em NeurIPS Workshop on Relational Representation Learning},
  2018.

\bibitem[SMM14]{smAIG}
K.~Sun and S.~Marchand-Maillet.
\newblock An information geometry of statistical manifold learning.
\newblock In {\em ICML 31}, volume~32 of {\em PMLR}, pages 1--9, 2014.

\bibitem[SN17]{snRFI}
K.~Sun and F.~Nielsen.
\newblock Relative {F}isher information and natural gradient for learning large
  modular models.
\newblock In {\em ICML 34}, volume~70 of {\em PMLR}, pages 3289--3298, 2017.

\bibitem[TSE{\etalchar{+}}19]{robustness_odds}
D.~Tsipras, S.~Santurkar, L.~Engstrom, A.~Turner, and A.~Madry.
\newblock Robustness may be at odds with accuracy.
\newblock In {\em ICLR}, 2019.

\bibitem[VCC{\etalchar{+}}18]{vcGAN}
P.~Veli{\v{c}}kovi{\'{c}}, G.~Cucurull, A.~Casanova, A.~Romero, P.~Li{\`{o}},
  and Y.~Bengio.
\newblock Graph attention networks.
\newblock {\em ICLR}, 2018.

\bibitem[VFH{\etalchar{+}}19]{velickovic2018deep}
P.~Veli{\v{c}}kovi{\'{c}}, W.~Fedus, W.~L. Hamilton, P.~Li\`o, Y.~Bengio, and
  R~D. Hjelm.
\newblock Deep graph infomax.
\newblock In {\em ICLR}, 2019.

\bibitem[WSZ{\etalchar{+}}19]{wsSGC}
F.~Wu, A.~Souza, T.~Zhang, C.~Fifty, T.~Yu, and K.~Weinberger.
\newblock Simplifying graph convolutional networks.
\newblock In {\em ICML}, volume~97 of {\em PMLR}, pages 6861--6871, 2019.

\bibitem[XBN{\etalchar{+}}15]{svm_adv2}
H.~Xiao, B.~Biggio, B.~Nelson, H.~Xiao, C.~Eckert, and F.~Roli.
\newblock Support vector machines under adversarial label contamination.
\newblock In {\em Neurocomputing}, volume 160, pages 53--62, 2015.

\bibitem[XLT{\etalchar{+}}18]{pmlr-v80-xu18c}
K.~Xu, C.~Li, Y.~Tian, T.~Sonobe, K.~Kawarabayashi, and S.~Jegelka.
\newblock Representation learning on graphs with jumping knowledge networks.
\newblock In {\em ICML 35}, volume~80 of {\em PMLR}, pages 5453--5462, 2018.

\bibitem[YCS16]{ycRSS}
Z.~Yang, W.~W. Cohen, and R.~Salakhutdinov.
\newblock Revisiting semi-supervised learning with graph embeddings.
\newblock In {\em ICML}, volume~48 of {\em PMLR}, pages 40--48, 2016.

\bibitem[YHC{\etalchar{+}}18]{yhGCN}
R.~Ying, R.~He, K.~Chen, P.~Eksombatchai, W.~L. Hamilton, and J.~Leskovec.
\newblock Graph convolutional neural networks for web-scale recommender
  systems.
\newblock In {\em KDD}, pages 974--983, 2018.

\bibitem[ZFY{\etalchar{+}}19]{zfTAA}
C.~Zhao, P.~T. Fletcher, M.~Yu, Y.~Peng, G.~Zhang, and C.~Shen.
\newblock The adversarial attack and detection under the {F}isher information
  metric.
\newblock In {\em AAAI}, 2019.

\bibitem[ZSDG18]{zsNNG}
G.~Zhang, S.~Sun, D.~Duvenaud, and R.~Grosse.
\newblock Noisy natural gradient as variational inference.
\newblock In {\em ICML}, volume~80 of {\em PMLR}, pages 5852--5861, 2018.

\end{thebibliography}

\ifdefined\arxiv
\bibliographystyle{alpha}
\onecolumn
\appendix
\section*{Supplementary Material of  ``Fisher-Bures Adversary Graph Convolutional Networks''}

\section{Proof of \Cref{thm:bm}}

By \cref{eq:bm}, we have
\begin{equation}\label{eq:bmrepeat}
ds^2 = \frac{1}{2}\sum_{j=1}^n \sum_{k=1}^n
\frac{(u_j^\top d\rho u_k)^2}{\lambda_j+\lambda_k}.
\end{equation}
We also have
\begin{equation*}
d\rho = \sum_{i=1}^n
\left[ d\lambda_i u_i u_i^\top
+  \lambda_i du_i u_i^\top
+ \lambda_i u_i du_i^\top
\right].
\end{equation*}
Because $\{u_i\}$ are orthonormal, we have
\begin{equation*}
\left(u_j^\top\sum_{i=1}^n
\left[ d\lambda_i u_i u_i^\top
+  \lambda_i du_i u_i^\top
+ \lambda_i u_i du_i^\top
\right]u_k\right)
=
d\lambda_j \delta_{jk}
+ \lambda_k u_j^\top du_k
+ \lambda_j du_j^\top u_k
\end{equation*}
Wrt the $\lambda$ parameters, we have
\begin{equation}
ds^2 = \frac{1}{2}\sum_{j} \frac{(d\lambda_j)^2}{2\lambda_j} = \frac{1}{4} \sum_{i} \frac{1}{\lambda_i} d\lambda_i^2
\end{equation}
The first term on the rhs is proved. Now we consider the $U$ parameters.
On the unitary group, we have $\forall{j},{k}$,
\begin{equation}
u_j^\top u_k = \mathrm{constant},
\end{equation}
therefore
\begin{equation}
d(u_j^\top u_k) = du_j^\top u_k + u_j du_k^\top = 0.
\end{equation}
Therefore
\begin{align}
(\lambda_k u_j^\top du_k + \lambda_j du_j^\top u_k)^2
&=
(\lambda_k u_j^\top du_k + \lambda_k  du_j^\top u_k + (\lambda_j-\lambda_k) du_j^\top u_k)^2\nonumber\\
&=
(\lambda_j-\lambda_k)^2 (du_j^\top u_k)^2.
\end{align}
Plugging back into \cref{eq:bmrepeat}, we get the Riemannian metric in
the $U$ coordinates. Notice that the cross terms
$d\lambda_i du_i$ are ignored.

\section{Proof of \cref{thm:bmtheta}}
The result is straightforward by plugging
\begin{equation}
d\lambda_i = \exp(\theta_i) d\theta_i
\end{equation}
into the statement of \cref{thm:bm}.

\section{Proof of \cref{thm:bmscale}}

We only need to prove the first part of \cref{thm:bmscale},
that leads to the second part.

By \cref{thm:bm}, we have
\begin{align}
\trace(\fim(u_i))
&= \frac{1}{2}
\trace\left(
\sum_{j=1}^n
\left(
\frac{(\lambda_i-\lambda_j)^2}{\lambda_i+\lambda_j}
u_j u_j^\top
\right)
\right)
\nonumber\\
&= \frac{1}{2}
\sum_{j=1}^n
\trace
\left(
\frac{(\lambda_i-\lambda_j)^2}{\lambda_i+\lambda_j}
u_j u_j^\top
\right)\nonumber\\
&=
\frac{1}{2}
\sum_{j=1}^n
\left(
\frac{(\lambda_i-\lambda_j)^2}{\lambda_i+\lambda_j}
\trace (u_j u_j^\top)
\right)\nonumber\\
&=
\frac{1}{2} \sum_{j=1}^n
\frac{(\lambda_i-\lambda_j)^2}{\lambda_i+\lambda_j}.
\end{align}
Because
\begin{equation}
\frac{\vert\lambda_i-\lambda_j\vert}{\lambda_i+\lambda_j} \le 1,
\end{equation}
we got a stronger result
\begin{equation}
\trace(\fim(u_i)) \le
\frac{1}{2} \sum_{j=1}^n \vert \lambda_i - \lambda_j \vert.
\end{equation}
Note that for density matrix the trace are normalized and we have
$0\le\lambda_i\le1$, Therefore
\begin{equation}
\trace(\fim(u_i)) \le
\frac{1}{2} \sum_{j=1}^n \vert \lambda_i - \lambda_j \vert
\le
\frac{1}{2} \sum_{j=1}^n \vert 0 - \lambda_j \vert
=
\frac{1}{2} \sum_{j=1}^n \lambda_j = \frac{1}{2}.
\end{equation}

\section{Proof of \cref{thm:lowerank}}

We first notice that $D_B$ is invariant to unitary transformations:
for any unitary $U$, we have
\begin{equation}
D_B(U\rho_1U^\top, U\rho_2U^\top)
=
D_B(\rho_1, \rho_2).
\end{equation}
Therefore
\begin{equation}
D_B(\rho, \rho_0) =
D_B(\rho, U\Lambda U^\top)
=
D_B(U^\top \rho U, \Lambda),
=
D_B(U^\top V R V^\top U, \Lambda)
\end{equation}
where $\Lambda=\diag(\lambda)$, and
$R=\diag(r_1,\cdots,r_n)$.
By Theorem 3~\cite{mmQSD},
the optimal $V^\star=U$ so that the
first density matrix on the rhs is diagonal,
and the optimal $R$ must have the same order as $\Lambda$.
The problem reduces to
\begin{equation}
\min 2( 1 - \sum_{i} \sqrt{r_i \lambda_i} )
\end{equation}
with respect to the constraints
\begin{align}
\forall{i}, r_i &\ge 0\\
\sum_{i} r_i &= 1\\
\text{$r$ has $k$ non-zero entries}
\end{align}
The optimal $r^\star$ must be composed of the
largest $k$ eigenvalues of the given density matrix, i.e.,
$\lambda_1$, $\lambda_k$ after re-scaling, that is,
\begin{equation}
\left\{
\begin{array}{lll}
r_i &= \gamma \lambda_i & \text{(if $i=1,\cdots{}k$)}\\
r_i &= 0 & \text{(otherwise)}
\end{array}
\right.
\end{equation}
We have
\begin{equation}
\sum_i r_i = \gamma \sum_{i}\lambda_i  = 1.
\end{equation}
Therefore
$\gamma = 1/\sum_{i}\lambda_i$.
Now we have both $R$ and $V$ and can express the optimal
low-rank projection, which is given by \cref{thm:lowerank}.

\section{Proof of \cref{thm:fim}}

By \cref{eq:gcn}, we have
\begin{equation}
dH^{l+1} =  \Sigma \circ (\tilde{A} dH^l W^l)
+ \Sigma \circ (\tilde{A} H^l dW^l)
+ \Sigma \circ (d \tilde{A} H^l W^l ).
\end{equation}
and
\begin{align}
&d\ell\nonumber\\
&= \trace( \frac{\partial\ell}{\partial H^{l+1}}^\top dH^{l+1})
\nonumber\\
&=
\trace\left( \frac{\partial\ell}{\partial H^{l+1}}^\top
\left(
\Sigma \circ (\tilde{A} dH^l W^l)
+ \Sigma \circ (\tilde{A} H^l dW^l)
+ \Sigma \circ (d \tilde{A} H^l W^l) \right)\right)\nonumber\\
&=
\trace\left( ( \frac{\partial\ell}{\partial H^{l+1}}^\top \circ \Sigma^\top )
(\tilde{A} dH^l W^l)\right)
+  \trace\left(
( \frac{\partial\ell}{\partial H^{l+1}}^\top \circ \Sigma^\top )
(d\tilde{A} H^l W^l) \right)\nonumber\\
&+
\trace\left(
( \frac{\partial\ell}{\partial H^{l+1}}^\top \circ \Sigma^\top )
(\tilde{A} H^l dW^l) \right)\nonumber\\
&=
\trace\left( W^l
(\frac{\partial\ell}{\partial H^{l+1}} \circ \Sigma)^\top
\tilde{A} dH^l \right)
+  \trace\left(
( \frac{\partial\ell}{\partial H^{l+1}} \circ \Sigma )^\top
\tilde{A} H^l dW^l \right)\nonumber\\
&+
 \trace\left(
 H^l W^l
( \frac{\partial\ell}{\partial H^{l+1}} \circ \Sigma )^\top
d \tilde{A} \right)
\end{align}
Therefore
\begin{align}
\frac{\partial\ell}{\partial{H}^{l}}
&=
\tilde{A} \left( \frac{\partial\ell}{\partial H^{l+1}}
\circ \Sigma \right) W^{l\top};\nonumber\\
\frac{\partial\ell}{\partial{W}^{l}}
&=
H^{l\top}
\tilde{A}
( \frac{\partial\ell}{\partial H^{l+1}} \circ \Sigma );
\nonumber\\
\frac{\partial\ell}{\partial{\tilde{A}}}
&=
H^l W^l
( \frac{\partial\ell}{\partial H^{l+1}} \circ \Sigma ).
\end{align}
Note only all layers contributes to the gradient \wrt $\tilde{A}$,
and the above expression has to be corrected accordingly.
Strictly speaking, this gradient has to be projected to be
symmetric based on the constraint of the $\tilde{A}$ matrix.

The stated results are straightforward from
the definition of the FIM (see \cite{aIGI}) in \cref{eq:fim},
and the above chain-rule equations.

\section{Learning Curves}

See \cref{fig:curves} for learning curves on the CiteSeer and PubMed datasets.
One can observe that the proposed FisherGCN and FisherGCN$^T$ have better
training and validation scores during learning. Their performance improvement
is more significant at earlier epochs. These curves are evaluated on 
the training and validation datasets. See \cref{tbl:results} for the
final scores on the testing datasets. See \cref{tbl:origresults} for
the testing scores using the Planetoid split~\cite{ycRSS}. Observe that
different split lead to a large variation of the testing scores.

\begin{figure}[t]
\centering
\begin{subfigure}[b]{.7\textwidth}
\includegraphics[width=\textwidth]{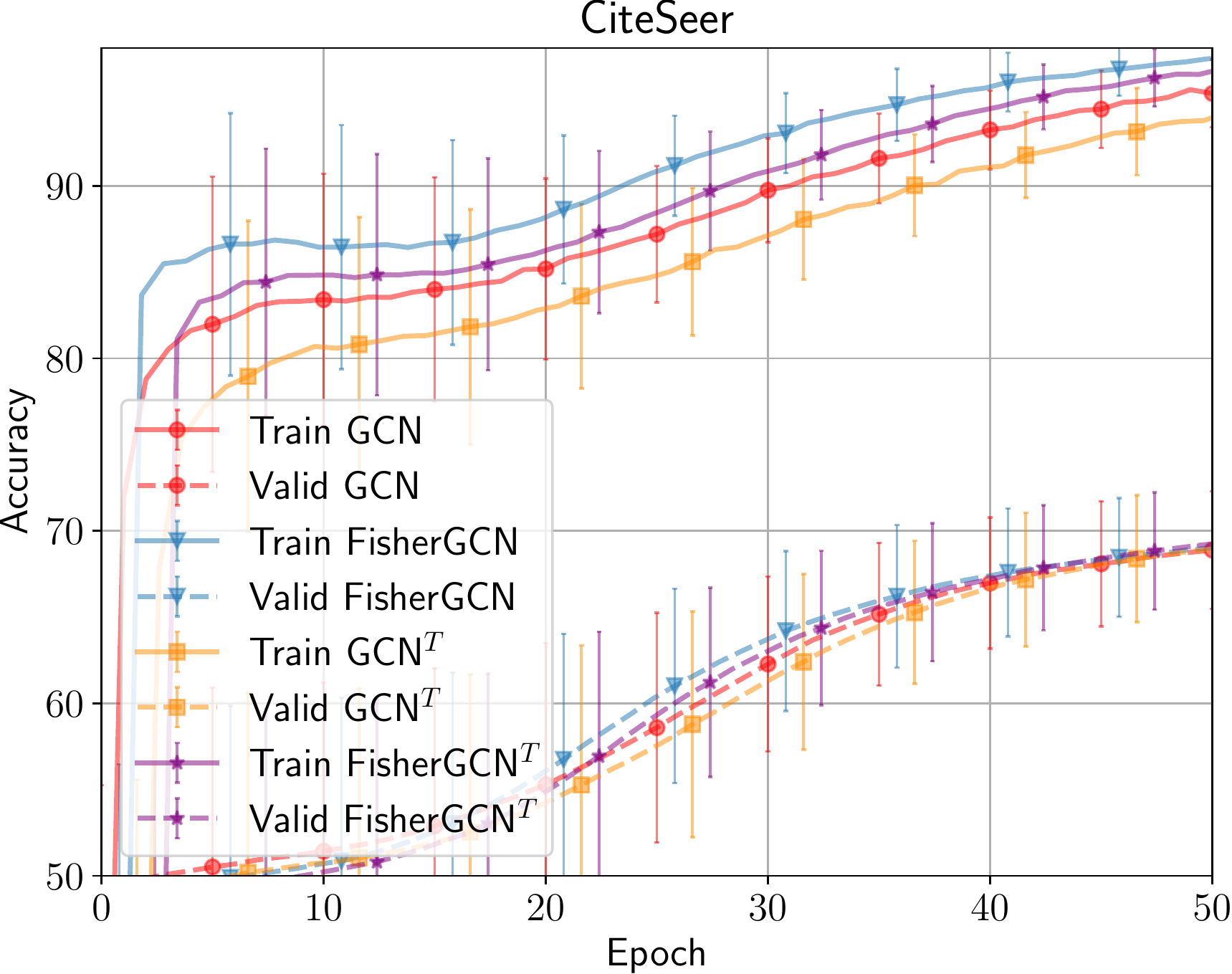}
\end{subfigure}
\begin{subfigure}[b]{.7\textwidth}
\includegraphics[width=\textwidth]{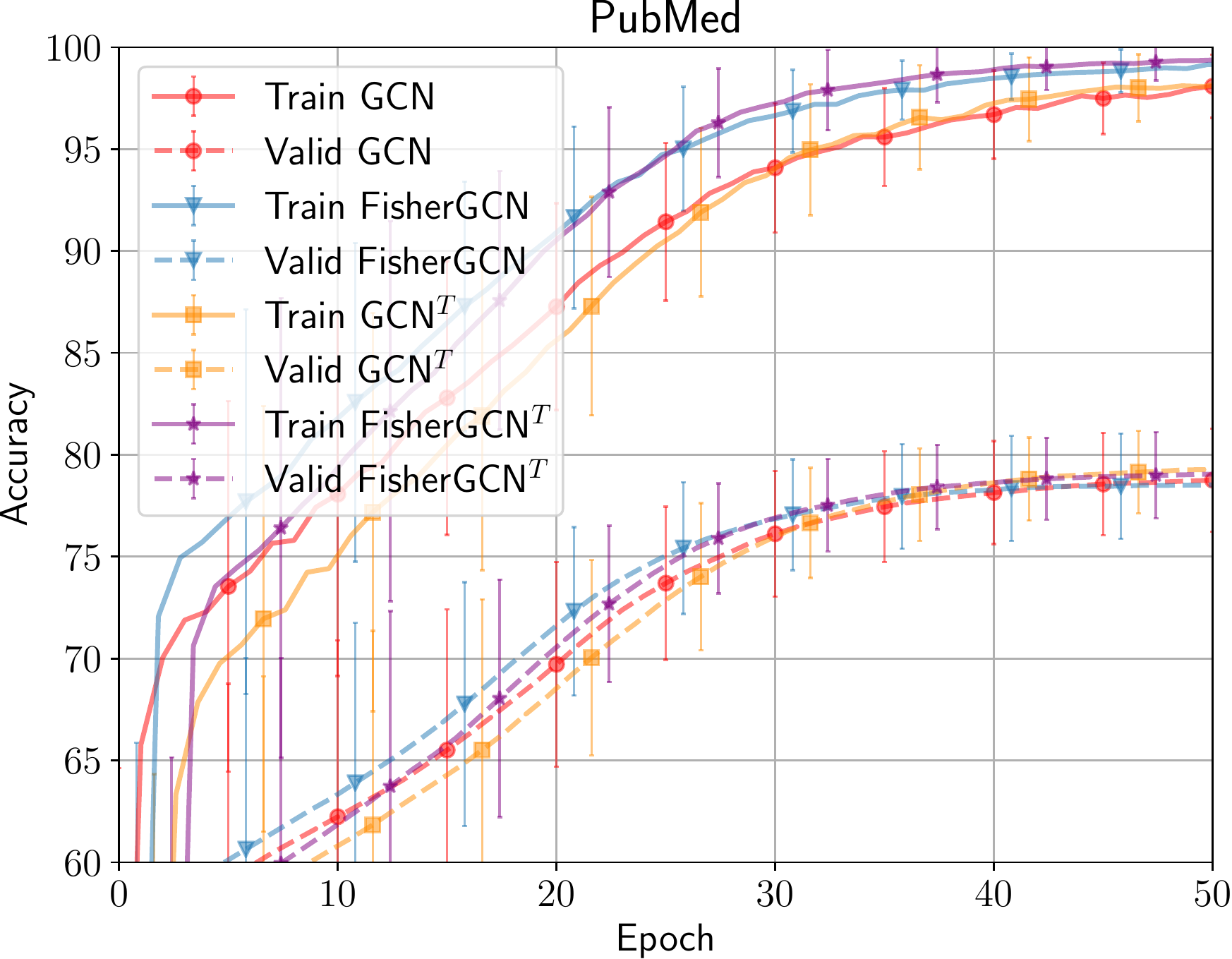}
\end{subfigure}
\caption{Learning curves (over 200 runs on 20 different splits of the training:validation:testing datasets) in accuracy on the CiteSeer and Pubmed datasets.\label{fig:curves}}
\end{figure}

\begin{table*}[t]
\caption{Testing loss and accuracy on the canonical split~\cite{ycRSS} using the 
same hyperparameters (learning rate 0.01; 64 hidden units; dropout rate 0.5; weight decay $5\times10^{-4}$).
The reported mean$\pm$std scores are based on 50 different initializations of the
neural network weights.\label{tbl:origresults}}
\begin{tabular}{c|ccc|ccc}
\hline
& \multicolumn{3}{|c|}{Testing Accuracy} & \multicolumn{3}{|c}{Testing Loss} \\
\hline & Cora & CiteSeer & PubMed & Cora & CiteSeer & PubMed \\
\hline
\hline GCN~\cite{kwSSC} & 81.5 & 70.3 & 79.0 & $-$ & $-$ & $-$\\
GCN              & $81.42\pm0.5$ & $70.62\pm0.5$ & $78.81\pm0.4$
                 & $1.07\pm0.01$ & $1.37\pm0.01$ & $0.74\pm0.01$ \\
FisherGCN        & $81.87\pm0.3$ & $70.92\pm0.3$ & $78.92\pm0.3$
                 & $1.06\pm0.00$ & $1.36\pm0.00$ & $0.73\pm0.00$ \\
GCN$^T$          & $81.88\pm0.4$ & $71.61\pm0.3$ & $79.11\pm0.4$
                 & $1.05\pm0.01$ & $1.33\pm0.01$ & $0.70\pm0.00$ \\
FisherGCN$^T$    & $82.20\pm0.3$ & $71.82\pm0.3$ & $79.05\pm0.2$
                 & $1.03\pm0.00$ & $1.32\pm0.00$ & $0.69\pm0.00$ \\
\hline
\end{tabular}
\begin{tabular}{ccc}
\hline
\end{tabular}
\end{table*}

\section{Experimental Settings}

As all our methods are different implementations of GCN,
we first tune the vanilla GCN on the Cora and CiteSeer datasets
based on random splits of the training:validation:testing datasets
over the following configuration grid:
\begin{itemize}
\item learning rate \{0.02, 0.01, 0.005, 0.001\};
\item Dropout rates \{0.5, 0.8\};
\item $L^2$ regularization strength \{0.002, 0.001, 0.0005\};
\item Number of layers 2;
\item Hidden layer dimensionality \{16,32,64\};
\end{itemize}
We try to select the best configuration as indicated in the caption of
\cref{tbl:results} based on the overall testing scores on these two datasets.
Notice that this ``best'' choice depends on the early stopping strategy
and the random splits used in the parameter searching process.
Then, we apply the exactly the same set of hyperparameters to all methods and datasets.

We set the maximum number of epochs to 500 and use the same early stopping strategy for all compared methods.
The learner is terminated is the 10-MA (moving average over the past 10 epochs)
validation loss turns larger than the 100-MA validation loss, and
the 10-MA validation accuracy turns smaller than the 100-MA validation accuracy.

\section{Proof of \cref{thm:embed}}

We denote the KL divergence as
\begin{equation*}
\mathcal{E}=\sum_{i}\sum_{j}w_{ij}\log\frac{w_{ij}}{p_{ij}}
= \sum_{i}\left[ \sum_{j}\left[ w_{ij}\log w_{ij} + w_{ij} D_{ij} \right]
+ \log\sum_{j}\exp(-D_{ij}) \right],
\end{equation*}
where $D_{ij}=\Vert{y}_i-{y}_j\Vert^2$. Therefore
\begin{align}\label{eq:derivative}
d\mathcal{E} &=
\sum_{i} \left[ \sum_{j} w_{ij} dD_{ij} + \frac{1}{Z_i} \sum_{j} \exp(-D_{ij}) (-dD_{ij}) \right]\nonumber\\
&=
\sum_{i} \sum_{j} \left( w_{ij} - p_{ij}(Y) \right) dD_{ij}.
\end{align}
As
\begin{equation*}
d D_{ij} = d \Vert{y}_i - {y}_j\Vert^2 = 2 ({y}_i - {y}_j )^\top (d{y}_i - d{y}_j ),
\end{equation*}
we have
\begin{align*}
d\mathcal{E} &=
\sum_{i} \sum_{j} \left( w_{ij} - p_{ij}(Y) \right) 2 (y_i - y_j )^\top (d{y}_i - d{y}_j )\nonumber\\
&=
2 \sum_{i} \sum_{j} \left( w_{ij} - p_{ij}(Y) \right) (y_i - y_j )^\top (d{y}_i - d{y}_j)\nonumber\\
&=
2 \trace\left( d{Y}^\top \diag\left( ({W} - P(Y)){1} \right) {Y} \right)\nonumber\\
&+ 2 \trace\left( d{Y}^\top \diag\left( ({W}^\top - {P}^\top(Y))1 \right) Y \right)\nonumber\\
&- 2 \trace\left( d{Y}^\top \left({W} - {P}(Y)\right) Y \right)\nonumber\\
&- 2 \trace\left( d{Y}^\top \left(W^\top - P^\top(Y)\right) Y \right).
\end{align*}
Therefore
\begin{align*}
\frac{\partial\mathcal{E}}{\partial{Y}}
=&\; 2 \diag\left(({W} - {P}({Y})){1}\right) {Y} + 2 \diag\left(({W}^\top - {P}^\top(Y)){1}\right) Y\nonumber\\
&- 2 \left(W - P(Y)\right) Y\nonumber\\
&- 2 \left(W^\top - P^\top(Y)\right) Y\nonumber\\
=&
4 L_{\mathrm{sym}}({W} - P(Y)) Y,
\end{align*}
where
${L}_{\mathrm{sym}}(W - P(Y))$ is the Laplacian matrix wrt the indefinite
weights $\frac{1}{2}\left[W + W^\top - P(Y) - P^\top(Y)\right]$.

By \cref{eq:derivative},
\begin{align}\label{eq:derivative2}
d^2 \mathcal{E}
= \sum_{i} \sum_{j} (w_{ij} - p_{ij}(Y)) d^2 D_{ij}
- \sum_{i} \sum_{j} d p_{ij}(Y) dD_{ij}.
\end{align}
By noticing
\begin{align*}
d^2 D_{ij}
&= 2 \left(dy_i - dy_j\right)^\top \left(dy_i - dy_j\right) = 2 \sum_{l} \left(dy_{il} - d{y}_{jl}\right)^2,
\end{align*}
the first term on the RHS of \cref{eq:derivative2} turns out to be
\begin{align}
d^2 \mathcal{E}_1 &=
2 \sum_{i} \sum_{j} \sum_{l} (w_{ij} - p_{ij}(Y)) \left(dy_{il} - d{y}_{jl}\right)^2.
\end{align}
Therefore,
\begin{align}
\frac{\partial^2 \mathcal{E}_1}{\partial{y}_{l} \partial{y}_{l}} &= 4 L_{\mathrm{sym}}(W - P(Y)).
\end{align}

The second term of \cref{eq:derivative2} yields
\begin{align}
d^2 \mathcal{E}_2
=&- \sum_{i} \sum_{j} d p_{ij}(Y) dD_{ij}\nonumber\\
=&
- \sum_{i} \sum_{j} d \left(\frac{1}{Z_i} \exp(-D_{ij}) \right) dD_{ij}\nonumber\\
=&
 \sum_{i} \sum_{j} \frac{1}{Z_i} \exp(-D_{ij}) \left(dD_{ij}\right)^2
-\sum_{i} \sum_{j} \frac{1}{Z_i^2} \exp(-D_{ij}) dD_{ij} \sum_{j} \exp(-D_{ij}) dD_{ij} \nonumber\\
=&
 \sum_{i} \left[\sum_{j} p_{ij}(Y) \left(dD_{ij}\right)^2 -\left(\sum_{j} p_{ij}(Y) dD_{ij}\right)^2 \right].
\end{align}
We have
\begin{align*}
(d D_{ij})^2 &= 4 \sum_{k} \sum_{l} (y_{ik}-y_{jk}) (y_{il}-y_{jl}) (d{y}_{ik}-d{y}_{jk}) (d{y}_{il}-d{y}_{jl}),
\end{align*}
and therefore
\begin{align}
d^2\mathcal{E}_2 =
& 4 \sum_{i}\sum_{j}\sum_{k}\sum_{l} p_{ij}(Y) (y_{ik}-y_{jk}) (y_{il}-y_{jl}) (d{y}_{ik}-d{y}_{jk}) (d{y}_{il}-d{y}_{jl})\nonumber\\
&-4 \sum_{i} \left( \sum_j p_{ij}(Y) \sum_{k} (y_{ik}-y_{jk}) (d{y}_{ik}-d{y}_{jk}) \right)^2\nonumber\\
=&
4 \sum_{i}\sum_{j}\sum_{k} p_{ij}(Y) (y_{ik}-y_{jk})^2 (d{y}_{ik}-d{y}_{jk})^2 \nonumber\\
&-4 \sum_{i} \sum_{k} \left( \sum_j p_{ij}(Y) (y_{ik}-y_{jk}) (d{y}_{ik}-d{y}_{jk}) \right)^2
\text{(ignoring all terms with $k\neq{l}$)}
\end{align}

For the first term, we have
\begin{align}
\frac{\partial^2 \mathcal{E}_{21}}{\partial y_{k} \partial y_{k}}
&=
8L_{\mathrm{sym}}( {P}\circ{D}^k ).
\end{align}
where $D_{ij}^k = (y_{ik} - y_{jk})^2$ means the pair-wise distance along the $k$'th dimension,
and ``$\circ$'' is elementwise product.

For the second term, we have
\begin{align}
\frac{\partial^2 \mathcal{E}_{22}}{\partial{y}_k \partial{y}_k^\top} = -4(B^k)^\top (B^k),
\end{align}
where
\begin{equation}
b_{ij}^k = \left\{
\begin{array}{ll}
\sum_{j} p_{ij}(Y)(y_{ik}-y_{jk}) & \text{if }i=j;\\
- p_{ij}(Y)(y_{ik}-y_{jk}) & \text{otherwise}.
\end{array}
\right.
\end{equation}
Putting everything together, we get
\begin{align}
\frac{\partial^2\mathcal{E}}{\partial{y}_k\partial{y}_k^\top}
&=
4 L_{\mathrm{sym}}(W-P(Y)) + 8L_{\mathrm{sym}}(P(Y)\circ{D}^k) - 4(B^k)^\top B^k.
\end{align}

\else
\bibliographystyle{plain}
\fi

\end{document}